\documentclass[runningheads]{llncs}
\usepackage{graphicx}
\usepackage{amssymb}
\usepackage{amsmath}
\usepackage[colorlinks,linkcolor=blue,anchorcolor=blue,urlcolor=blue,citecolor=blue,CJKbookmarks=True]{hyperref} 
\usepackage{float}
\usepackage{amssymb}
\usepackage{booktabs}
\usepackage{threeparttable}
\usepackage{upgreek}
\usepackage{multirow}
\usepackage{marvosym}
\usepackage{orcidlink}
\begin{document}
\title{Image Gradient-Aided Photometric Stereo Network}
\titlerunning{IGA-PSN}
%
\author{Kaixuan Wang\inst{1} \and Lin Qi\inst{(\textrm{\Letter})1} \and Shiyu Qin\inst{1} \and Kai Luo\inst{1}\orcidlink{0009-0001-3829-8426} \and Yakun Ju\inst{2}\orcidlink{0000-0003-4065-4108}\and Xia Li\inst{1} \and Junyu Dong\inst{1}\orcidlink{0000-0001-7012-2087}}
\authorrunning{K. Wang et al.}
%
\institute{Department of Computer Science and Technology, Ocean University of China, Qingdao, China \\ \email{qilin@ouc.edu.com} \and School of Electrical and Electronic Engineering, Nanyang Technological University, Singapore, Singapore}
\maketitle              
\begin{abstract}
Photometric stereo (PS) endeavors to ascertain surface normals using shading clues from photometric images under various illuminations. Recent deep learning-based PS methods often overlook the complexity of object surfaces. These neural network models, which exclusively rely on photometric images for training, often produce blurred results in high-frequency regions characterized by local discontinuities, such as wrinkles and edges with significant gradient changes. To address this, we propose the Image Gradient-Aided Photometric Stereo Network (IGA-PSN), a dual-branch framework extracting features from both photometric images and their gradients. Furthermore, we incorporate an hourglass regression network along with supervision to regularize normal regression. Experiments on DiLiGenT benchmarks show that IGA-PSN outperforms previous methods in surface normal estimation, achieving a mean angular error of 6.46 while preserving textures and geometric shapes in complex regions.
\end{abstract}
\section{Introduction}
Photometric Stereo \cite{1_woodham1980photometric} aims to recover the surface normals of an object by utilizing shading clues extracted from images acquired under diverse illuminations. Distinguished from other stereo vision methods, PS obviates the necessity for intricate feature point matching procedures, particularly efficacious on textureless surfaces—a capability that eludes other texture-based stereo-vision techniques. 
\begin{figure*}[t]
\centering
\includegraphics[scale=0.7]{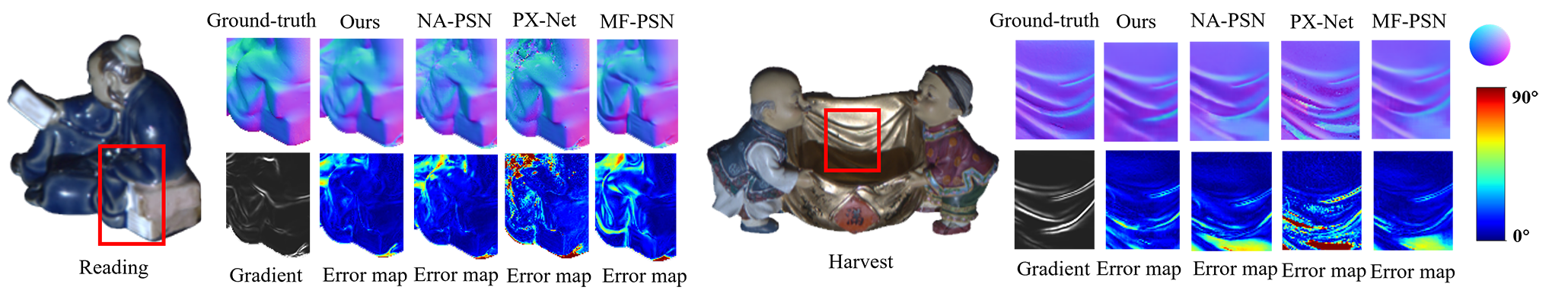}
\caption{An example of the errors in complex-structured regions. We visualized the gradient map of the high-frequency region. The results compare our method with the NA-PSN \cite{26_ju2022normattention}, PX-Net \cite{28_logothetis2021px} and MF-PSN \cite{16_liu2022deep}.}
\label{fig:img1}
\end{figure*}

Recently, learning-based PS methods~\cite{3_santo2017deep,4_ikehata2018cnn,12_chen2018ps} directly learn the mapping from observations to surface normals, offering enhanced convenience and superior robustness in reconstructing surface normals of non-Lambertian objects compared to conventional approaches. However, as Fig.\ref{fig:img1}, these studies struggle to handle intricate regions, such as those characterized by complex structures like wrinkles and edges, resulting in negative impact of normal estimation. 

Existing PS networks \cite{3_santo2017deep,12_chen2018ps,27_ju2023estimating,38_yang2023accurate} indiscriminately capture information across diverse frequencies, a method that can hinder the network's ability to effectively distinguish between high- and low-frequency information, thus affecting performance \cite{42_ju2023deep}. High-frequency regions, typically characterized by locally discontinuous features \cite{26_ju2022normattention}, require greater attention to sharpness and structure. We discover that complex regions can be easily delineated by gradients, as illustrated in Fig.\ref{fig:img1}. Consequently, we propose IGA-PSN, a multi-path dual-branch parallel network designed to handle both high- low-frequency regions by introducing a structural gradient extractor to delve deeper into gradient information from input images, guiding high-quality surface normal restoration. By using gradient cues to enhance common features and highlight gradients' auxiliary role in feature extraction, we develope an attention feature fusion module. This module empowers the network focus on specific local regions and adaptively aggregates model-specific features by adjusting weights in a learnable manner, optimizing feature extraction. Furthermore, previous works conventionally leverage cosine similarity in loss functions, which only considers the average angular difference, overlooking surface regions rich in details \cite{27_ju2023estimating}. While effective in recovering low-frequency global structures \cite{41_Isola_2017_CVPR,42_ju2023deep}, these methods often fail to preserve finer details, resulting in blurred and overly smoothed outputs. To address this, our method adds a gradient error loss of surface normals to the overall loss function. Here, gradients calculate the variation in surface normals between adjacent pixels, making gradient loss crucial for focusing on structural differences and ensuring clear and accurate surface normal recovery. Additionally, we design an hourglass regressor module with multi-level supervision to iteratively engage in normal regression in a top-down fashion, maximizing global information use \cite{23_chang2018pyramid}.

The effectiveness of the model is clearly proved by ablation studies, while benchmark comparisons on DiLiGenT Benchmark \cite{11_shi2016benchmark} and DiLiGenT-$\mathrm{\Pi}$ Dataset \cite{25_wang2023diligent} demonstrate the superior performance of our work.
\section{Related work}
The fundamental image formation equation governing PS method, delineating the relationship between surface normals $\mathbf{n} \in \mathbb{R}^3$ and pixel-wise observation $i$, is expressed as:
\begin{equation}
    \begin{aligned}
    i=\rho(\mathbf{n}, \mathbf{l})\max(\mathbf{n}^{\rm{T}}\mathbf{l},0)+\varepsilon  \\
    \end{aligned}
    \label{eq:eq1}
\end{equation}
where \(\rho\) denotes the general BRDF, a graphics parameter representing how light reflects off a surface from a specified incident direction \cite{13_wann2023practical}, and \(\mathbf{l}\) indicates the illumination with direction \(\mathbf{l}\). The term \(max(\mathbf{n}^T\mathbf{l}, 0)\) captures attached shadows, while \(\varepsilon\) represents noise sources such as inter-reflections and cast shadows. Traditional methods can be divided into outlier removal-based approaches \cite{7_holroyd2008photometric,18_ikehata2012robust,2_shi2013bi} and those modeling complex reflectance \cite{6_mukaigawa2007analysis,22_tozza2016direct,8_wu2009photometric}, but these models are only accurate for limited materials and suffer from unstable optimization.

In learning-based PS, \cite{3_santo2017deep} was the first to use deep neural networks to predict surface normals from PS images, but it required a fixed array of ordered light directions. Later work addressed this at full-pixel and per-pixel levels. All-pixel methods \cite{12_chen2018ps,16_liu2022deep,26_ju2022normattention} used max-pooling to aggregate features and learn pixel intensity variations. Per-pixel methods \cite{4_ikehata2018cnn,28_logothetis2021px,31_ikehata2022ps} used observation maps to learn intensity variations at the same pixel position across images. \cite{28_logothetis2021px} combined these approaches with a graph-based method to explore inter-image and intra-image information. However, current approaches often focus on designing deep normal regression networks or loss functions \cite{12_chen2018ps,26_ju2022normattention,31_ikehata2022ps}, frequently overlooking the importance of learning across different frequencies during image feature extraction.
\section{Methods}
In this section, we present our proposed PS network, IGA-PSN, designed to achieve high-fidelity recovery of surface normals especially in high-frequency regions. First, we introduce two preparatory operations for our PS network.
\subsection{Preliminaries}
\textbf{Data normalization:} Distinguishing between high-frequency regions caused by structural or textural factors is of paramount importance in observations, especially in handling real-word objects with abrupt color changes induced by surfaces with spatially varying BRDFs. Therefore, the adoption of the data normalization strategy \cite{15_chen2020deep} becomes necessary:
\begin{equation}
    \begin{aligned}
    i_j'=\frac{i_j}{\sqrt{i_1^2+i_2^2+\cdots +i_n^2}},\quad j\in\left \{ 1,2,\cdots ,n \right \}
    \end{aligned}
    \label{eq:eq2}
\end{equation}
where $i_1\cdots i_n$ indicate the pixel intensities of the same position in 1-st$\cdots$n-th images, $i_j'$ denotes the normalized pixel value of $i_j$.

Assuming Lambertian reflection, Eq. \ref{eq:eq1} simplifies to $i_j = \rho \max(\mathbf{n}^{\rm{T}}\mathbf{l},0)$. Substituting Eq.~\ref{eq:eq2} helps mitigate spatial BRDF variability from $\rho$. For non-Lambertian surfaces under directional lighting, low-frequency regions resemble Lambertian surfaces, while feature fusion disregards specular highlights and shadows, retaining salient features \cite{15_chen2020deep}. Thus, normalization can be applied to non-Lambertian surfaces to handle spatially varying BRDFs, especially in regions with complex structures or textures.

\textbf{Light embedding:} Each light direction, a three-dimensional (3D) vector, is replicated to form a three-channel tensor $\mathbb{R}^{H\times W\times3}$ with the same spatial dimensions as the input images. This is concatenated with the normalized data to create a $\mathbb{R}^{H\times W\times6}$ tensor as the network input, ensuring effective use of lighting information during surface normal recovery.
\begin{figure}[t]
\centering
\includegraphics[scale=0.38]{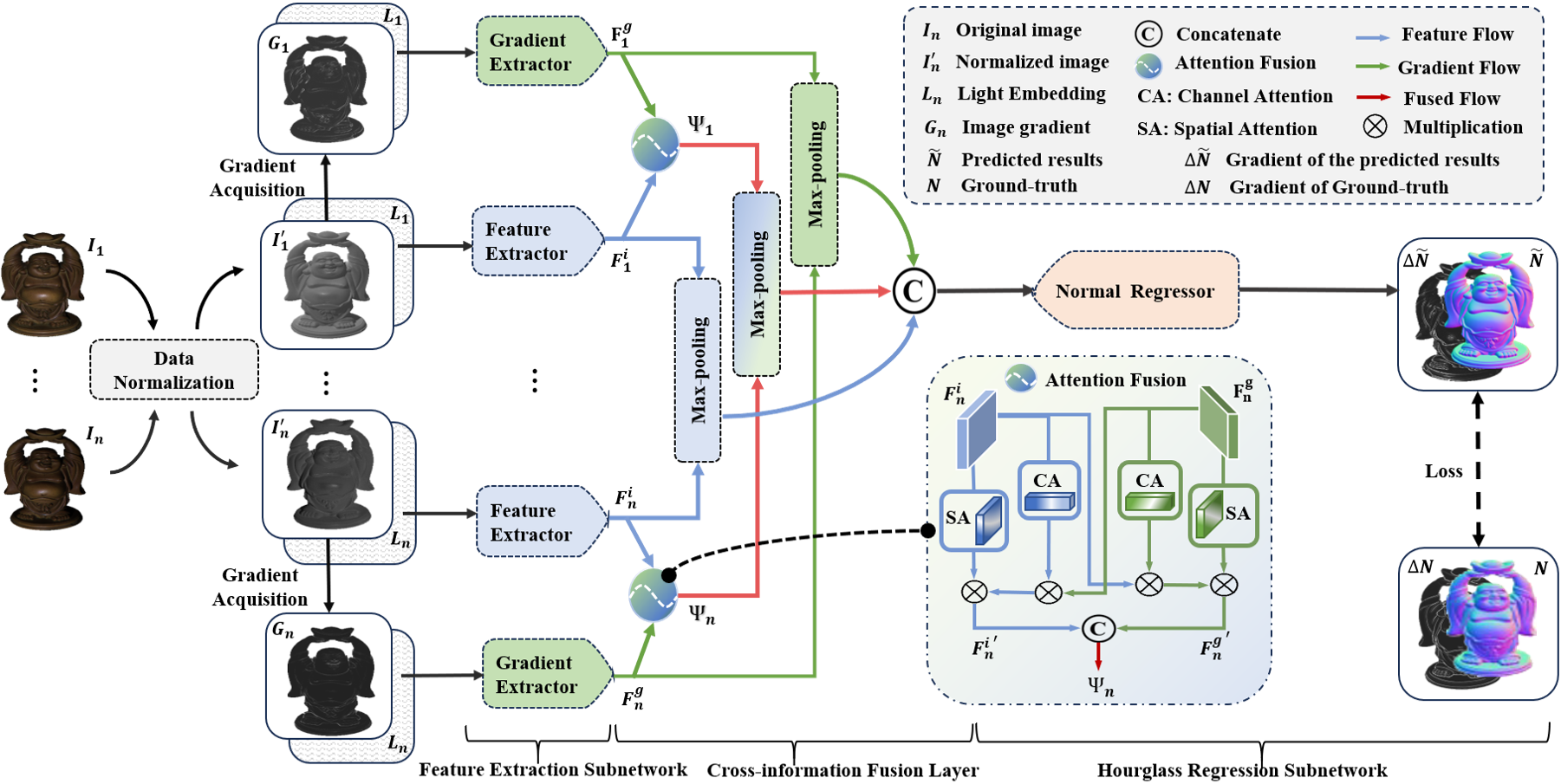}
\caption{The overview of IGA-PSN, which is composed of a shared-weight feature extraction subnetwork, cross-information fusion layer, and hourglass normal regression subnetwork.}
\label{fig:img2}
\end{figure}
\subsection{The proposed framework: IGA-PSN}
\begin{figure}[t]
\centering
\includegraphics[scale=0.6]{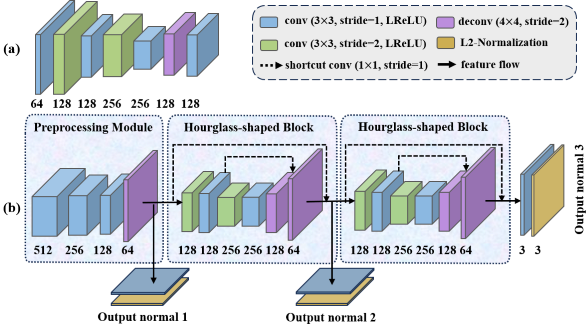}
\caption{Network details of feature/gradient extractor and normal regression subnetwork. The numbers below features indicate the dimension of the feature channel.}
\label{fig:img3}
\end{figure}

Our model, shown in Fig. \ref{fig:img2},  comprises three main components: a shared-weight feature extraction subnetwork, a cross-information fusion layer, and an hourglass normal regression subnetwork. Both the feature extraction and normal regression subnetworks, illustrated in Fig. \ref{fig:img3}, use fully convolutional layers for adaptability to varying input image sizes.

\noindent\textbf{The feature extraction stage}  is split into two branches to effectively handle both high- and low-frequency regions one extracts features from normalized images (feature extractor), and the other derives gradient feature maps (gradient extractor). Image gradients, indicating intensity or color changes between adjacent positions, are useful in tasks like super-resolution \cite{19_ma2020structure} and edge detection \cite{21_su2021pixel}. High-frequency regions with local discontinuities require attention to sharpness and structural details, which can be effectively captured by image gradients (\ref{fig:img1}). When combined with original normalized features, gradient features can act as an attention mechanism, enhancing normal estimation quality. We intentionally disregard the orientation (phase) of the gradient and use the magnitude for its effectiveness in revealing intricate textures and fine details in local regions. To reduce computational burden, we use the sum of the absolute values of the two gradient components instead of square or square-root operations in calculating gradient magnitude \cite{43_book}.

For the normalized observation \(i'\) at coordinates \((x, y)\), the simplified gradient map \(G\in \mathbb{R}^{H\times W\times3}\) is computed as follows: 
\begin{equation}
    \begin{aligned}
    G_{i'_{(x,y)}}& = \left\|\frac{i'(x+1,y)-i'(x-1,y)}{2}\right\|_{1}+\left\|\frac{i'(x,y+1)-i'(x,y-1)}{2}\right\|_{1} \\
    \end{aligned}
    \label{eq:eq3}
\end{equation}

As most regions of the gradient map are close to zero~\cite{19_ma2020structure}, convolutional neural networks can effectively focus on the spatial relationships of contours, facilitating the capture of structural dependencies. Additionally, textures, defined as patterns of local structures in images, are crucial for identifying different materials and surfaces. 
The normalized and gradient information will be respectively put into feature extractor and gradient extractor. And the features obtained through these two extraction processes are denoted as \(F^g\in \mathbb{R}^{\frac{1}{2} H\times\frac{1}{2} W\times128}\) for the gradient feature and \(F^i\in \mathbb{R}^{\frac{1}{2} H\times\frac{1}{2} W\times128}\) for the normalized image feature.

\noindent\textbf{Cross-information fusion layers:} Inspired by \cite{24_woo2018cbam}, we introduce an attention fusion module that integrates features from the normalized image branch (feature extractor result) and gradient branch (gradient extractor result). This module uses channel attention \(M_c\in \mathbb{R}^{C'\times1\times1}\) and spatial attention \(M_s\in \mathbb{R}^{1\times h\times w}\) to identify important channels and regions. An element-wise multiplication between the image and the gradient branches extracts complementary information for feature enhancement, as defined below:
\begin{equation}
    \begin{aligned}
    &F_j^{g'}=M_{c}(F^{g}_j)\otimes F^{i}_j \otimes M_{s}(F^{g}_j), \\
    &F_j^{i'}=M_{c}(F^{i}_j)\otimes F^{g}_j \otimes M_{s}(F^{i}_j) , \\
    &\Psi _j=\operatorname{Concat}(F_j^{g'},F_j^{i'})
    \end{aligned}
    \label{eq:eq4}
\end{equation}
where \(\otimes\) denotes element-wise multiplication, \(Concat(\cdot)\) signifies the concatenate operation, \(F^{g}_j\) and \(F^{i}_j\) is the output of the \(j\)-th gradient extractor and feature extractor, and \(\Psi_j\) is the fusion result of \(F^{g}_j\) and \(F^{i}_j\). This attention fusion bi-directionally connects the image and the gradient branches, focusing adaptively on distinct feature hierarchies and highlighting their spatial co-occurrence or correlation. Its efficacy is supported by ablation studies.

The final step involves feeding \(\Psi_j\), \(F^i_j\), and \(F^g_j\) into a max-pooling operation to extract salient features, represented by the following equation:
\begin{equation}
    \begin{aligned}
    \Gamma_{max}= Concat\{&Maxpool(\Psi_1,\Psi_2,\cdots,\Psi_n), \\ 
    &Maxpool(F_1^{g},F_2^{g},\cdots,F_n^{g}), \\ 
    &Maxpool(F_1^{i},F_2^{i},\cdots,F_n^{i})\}
    \end{aligned}
    \label{eq:eq5}
\end{equation}
where \(Maxpool(\cdot)\) denotes max-pooling, \(Concat(\cdot)\) denotes concatenation, and \(\Gamma _{max}\) represents the result of cross-information fusion layers. The outcomes are concatenated and then fed into a normal regressor for estimation.

\noindent\textbf{Regression subnetwork:} 
We introduce a novel hourglass normal regressor (Fig. \ref{fig:img3}) for surface normals regression. It includes a preprocessing module with four regular convolutions (LeakyReLU activation) and two transposed convolutions to up-sample fused feature maps; two stacked hourglass encoder-decoder modules used to capture contextual information through top-down/ bottom-up processing and intermediate supervision. This enhances learning by propagating supervision through layers. Both the preprocessing module and the hourglass blocks end with an L2 normalization layer, generating unit normal maps.

The regression subnetwork produces three levels of outputs and corresponding losses ($Loss_1$, $Loss_2$, and $Loss_3$), defined by Eq. \ref{eq:eq6}. During the training phase, the total loss is computed as the weighted sum of these losses. During testing, only the normal map from the last output is used for estimation. This approach ensures effective surface normal regularization, emphasizing contextual information capture and learning capabilities.

IGA-PSN optimizes its learnable parameters by minimizing the  loss function:
\begin{equation}
    \begin{aligned}
        &Loss=\sum_{k=1}^{3} \omega_k\cdot Loss_k ,\quad Loss_k=L_{A_k}+\mu L_{G_k} \\
    \end{aligned}
    \label{eq:eq6}
\end{equation}
where \(k\in \{1,2,3\}\) denotes the loss level, \(Loss_k\) and \(\omega_k\) indicate the loss and its weight (\(\omega_1=0.5\), \(\omega_2=0.7\), \(\omega_3=1.0\)). Besides the cosine similarity loss \(L_{A_k}\), we introduce a gradient difference loss \(L_{G_k}\) in this work, defining the gradient loss between ground-truth and estimated surface normals at the k-th output level (with \(\mu\) empirically set to 0.05, yielding the best results among experiments with values from 0.01 to 0.5). \(L_{A_k}\) and \(L_{G_k}\) are specifically defined as:
\begin{equation}
\begin{aligned}
%
L_{A_k}\!=\!\frac{1}{H W}\sum_{p}\left(1-\mathbf{n_{k}^{p}} \cdot \mathbf{\tilde{n}_{k}^p}\right),\quad L_{G_k}=\!\frac{1}{H W}\!\sum_{p}\left\| g(\mathbf{n_{k}^{p}})-g(\mathbf{\tilde{n}_{k}^p})\right\|_2
\end{aligned}
\label{eq:eq8}
\end{equation}
where $H$ and $W$ denote the dimension of the outputs, $\mathbf{n_{k}^{p}}$ and $\mathbf{\tilde{n}_{k}^p}$ denote the ground-truth and predicted surface normals at position $p(x,y)$ in the k-th output level. We define the gradient $g(\mathbf{n_{k}^{p}})$ like the format of Eq. \ref{eq:eq3} as follows:
\begin{equation}
g(\mathbf{n_{k}^{p}})=\left\|\frac{\mathbf{n_{k}^{p(x+1,y)} }-\mathbf{n_{k}^{p(x-1,y)}}}{2} \right\|_1+\left\|\frac{\mathbf{n_{k}^{p(x,y+1)}}-\mathbf{n_{k}^{p(x,y-1)} }}{2}\right\|_1
\label{eq:eq9}
\end{equation}

Compared to cosine similarity loss, gradient-based objectives enhance the network's focus on geometric structures. The gradient loss helps maintain the sharpness of high-curvature or discontinuous curved surfaces, avoiding the blurring of these high-frequency regions~\cite{19_ma2020structure}. And the gradient space constraint provides additional supervision, contributing to improved normal estimation. 

\section{Experiment}
In this section, we show experimental results as well as analysis of the our network. We employ mean angular error (MAE) in degrees to evaluate the accuracy of predicted surface normals by measuring the angular disparity between the predicted values and the ground-truth at each pixel:
\begin{equation}
    \begin{aligned}
    MAE=\frac{1}{HW} \sum_{p}\arccos(\mathbf{n_{p}} \cdot \mathbf{\tilde{n}_{p}}) \\
    \end{aligned}
    \label{eq:eq7}
\end{equation}
where \(HW\) represents the resolution of the output, and $\mathbf{n_{p}}$ and $\mathbf{\tilde{n}_{p}}$ denote the ground-truth and predicted surface normals at pixel \(p\), respectively. In addition to MAE, we introduce two supplementary metrics, \(err^{15^\circ}\) and \(err^{30^\circ}\), indicating the proportion of surface normals with angular deviations below \(15^\circ\) and \(30^\circ\), respectively. These metrics assess the model's performance in high-frequency regions with intricate surface details, where errors tend to be higher. \(err^{15^\circ}\) and \(err^{30^\circ}\) enhance comparative analysis, revealing the model's proficiency in recovering complex surface regions.

We apply Blobby \cite{34_johnson2011shape}, Sculpture dataset \cite{35_wiles2017silnet} for training, and DiLiGenT benchmark \cite{11_shi2016benchmark}, DiLiGenT-$\mathrm{\Pi}$ \cite{25_wang2023diligent} for evaluating. The training process was conducted on a single NVIDIA GeForce RTX 3090 24GB GPU for around 12 hours.
\subsection{Ablation study}
\textbf{Effectiveness of gradient extractor:} 
The outcomes from ID(1) to ID(3) underscore the importance of discriminating between the normalized image and the gradient map during processing. As ID(1), directly inputting both into the feature extractor fails to improve model performance. While handled separately (ID(3)), the introduction of a gradient extractor significantly reduced the MAE to 7.21. Despite this improvement, there is minimal impact on $err^{15^\circ}$, and $err^{30^\circ}$ even decreased.  This suggests that the indiscriminate use of gradient information without specific strategies may not lead to significant enhancements.
Analysis of ID(2) shows that relying solely on gradient information does not improve network accuracy.  A key reason is the prevalence of zero values in the gradient map, limiting its ability to capture features in flat regions.  While gradient information complements feature extraction, its exclusive use is insufficient for enhancing performance.

\begin{table}[htp]
\renewcommand{\arraystretch}{0.8}
\setlength\tabcolsep{4.7pt}
\caption{Ablation results on DiLiGenT benchmark \cite{11_shi2016benchmark}, in terms of MAE, $err^{15^\circ}$ and $err^{30^\circ}$. NI: Normalized Image; FE: Feature Extractor; GI: Gradient Information; GE: Gradient Extractor; AF: Attention Feature Fusion Module; $L_A$: Cosine Similarity Loss; $L_G$: Gradient Loss; HM: Hourglass regressor Module in normal regression network.}
\centering
\begin{tabular}{|c|cccccccc|ccc|}
\hline
ID & NI & FE & GI & GE & AF & $L_G$ & $L_A$ & HM & MAE ↓ & $err^{15^\circ}$ ↑& $err^{30^\circ}$↑\\
\hline
(0) &  $\checkmark$&  $\checkmark$& - & - & - & - &  $\checkmark$& - & 7.39 & 90.4\% & 97.4\% \\
(1) &  $\checkmark$&  $\checkmark$&  $\checkmark$& - & - & - &  $\checkmark$& - & 7.46 & 90.3\% & 97.2\% \\
(2) & - & - &  $\checkmark$&  $\checkmark$& - & - &  $\checkmark$& - & 26.10 & 45.5\% & 69.3\% \\
(3) &  $\checkmark$&  $\checkmark$&  $\checkmark$&  $\checkmark$& - & - &  $\checkmark$& - & 7.21 & 90.5\% & 97.0\% \\
(4) &  $\checkmark$&  $\checkmark$&  $\checkmark$&  $\checkmark$&  $\checkmark$& - &  $\checkmark$& - & 6.99 & 90.8\% & 97.5\% \\
(5) &  $\checkmark$&  $\checkmark$&  $\checkmark$&  $\checkmark$&  $\checkmark$&  $\checkmark$& - & - & 43.77 & 27.3\% & 35.0\% \\
(6) &  $\checkmark$&  $\checkmark$&  $\checkmark$&  $\checkmark$&  $\checkmark$&  $\checkmark$&  $\checkmark$& - & 6.82 & 91.1\% & 97.5\% \\
(7) &  $\checkmark$&  $\checkmark$&  $\checkmark$&  $\checkmark$&  $\checkmark$&  $\checkmark$&  $\checkmark$&  $\checkmark$& \textbf{6.68} & \textbf{91.8\%} & \textbf{98.0\%} \\
\hline
\end{tabular}
\label{tab:tab1}
\end{table}

\noindent\textbf{Effectiveness of the fusion approach:} 
ID(3), ID(4) validate the efficacy of the cross-information fusion layer. In ID(4), with the attention feature fusion module, improvements in MAE, \(err^{15^\circ}\), and \(err^{30^\circ}\) are observed. Cross-attention fusion is compared to vanilla CBAM \cite{24_woo2018cbam} in Table \ref{tab:tab2}, showing superior performance across all metrics. This enhances learning in high-frequency and smooth surfaces, emphasizing bidirectional alignment in attention feature fusion between the normalized image and gradient map.

\noindent\textbf{Choice of loss function:}
Subsequently, we assess the impact of incorporating gradient error loss. In Table \ref{tab:tab1}, from ID(4) to ID(6), it's evident that relying only on gradient error loss can prevent network convergence. This is due to the gradient loss focusing solely on variations in surface normal values between neighboring pixels, neglecting changes in angles between surface normals. Simultaneously considering both aspects in ID(6) results in metric improvements.
\begin{figure}[hbt]
\centering
\includegraphics[scale=0.45]{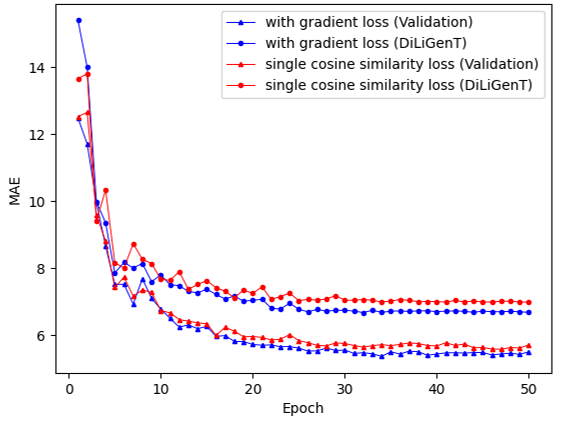}
\caption{Comparing the convergence of the models. The blue line represents our IGA-PSN, while the red line corresponds to a model using a single cosine similarity loss. Both models are trained with the same architecture over 50 epochs. The model optimized with both two loss functions shows a lower convergence error than the model using only cosine similarity loss. This shows the effectiveness of adding gradient error loss.}
\label{fig:img4}
\end{figure}
\begin{table}[hbt]
\renewcommand{\arraystretch}{0.82}
\setlength\tabcolsep{8pt}
    \centering
    \caption{Comparative experiments involving attention feature fusion mechanisms at the fusion stage. The \textbf{bold} values indicate the best results.}
    \begin{tabular}{|c|c|ccc|}
    \hline
        ID & Fusion Module & MAE ↓ & $err^{15^\circ}$ ↑& $err^{30^\circ}$↑ \\
    \hline
        (8) & using CBAM module \cite{24_woo2018cbam} & 7.68 & 88.9\% & 96.8\% \\
        (9) & using attention fusion module & \textbf{6.68} & \textbf{91.8\%} & \textbf{98.0\%} \\     				
    \hline
    \end{tabular}
    \label{tab:tab2}
\end{table}
To visualize convergence behavior, we compare results on the validation set and DiLiGenT benchmark \cite{11_shi2016benchmark} in Fig. \ref{fig:img4}. The model optimized using gradient error loss (blue line) is compared with one optimized solely using cosine similarity loss (red line), showing average MAE over 50 epochs. The model optimized with both cosine similarity and gradient loss achieves lower convergence error compared to the model with only cosine similarity loss, as demonstrated by the 34th epoch. This supports the effectiveness of incorporating gradient error loss.
\begin{figure}[hbt]
\centering
\includegraphics[scale=0.38]{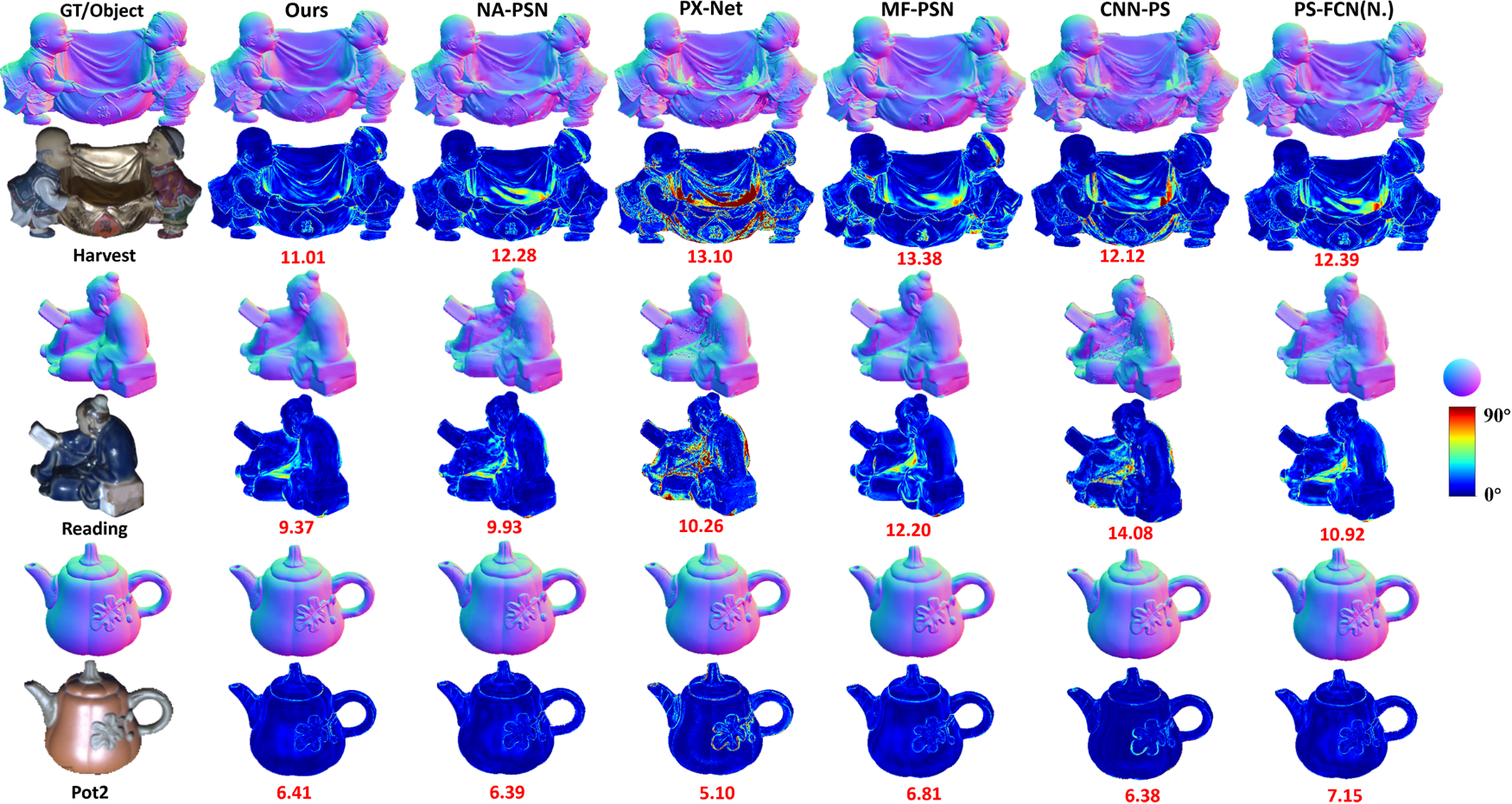}
\caption{Quantitative results on objects ``Harvest",``Reading" and ``Pot2" on the DiLiGenT benchmark \cite{11_shi2016benchmark} with 96 input images. Numbers below the normal map are the MAE in degrees. Compared with NA-PSN \cite{26_ju2022normattention}, PX-Net \cite{28_logothetis2021px}, MF-PSN \cite{16_liu2022deep}, CNN-PS \cite{4_ikehata2018cnn} and PS-FCN(N.) \cite{15_chen2020deep}, our model achieves the best or sub-optimal results.}
\label{fig:img5}
\end{figure}
\begin{table}[hbt]
\renewcommand{\arraystretch}{1}
\setlength\tabcolsep{0.78pt}
\caption{Comparisons on DiLiGenT main benchmark \cite{11_shi2016benchmark}. The \textbf{bold} values indicate the best results, while the \underline{underlined} values represent the second-best results.}
\begin{tabular}{|c|cccccccccc|c|}
\hline
Method & Ball & Bear & Buddha & Cat & Cow & Goblet & Harvest & Pot1 & Pot2 & Reading & Avg. \\
\hline
L2\cite{1_woodham1980photometric} & 4.10 & 8.39 & 14.92 & 8.00 & 25.60 & 18.50 & 30.62 & 8.89 & 14.65 & 19.80 & 15.39 \\
IRPS\cite{17_taniai2018neural} & \textbf{1.47} & 5.79 & 10.36 & 5.44 & 6.32 & 11.47 & 22.59 & 6.09 & 7.76 & 11.03 & 8.83\\
PS-FCN(N.)\cite{15_chen2020deep} & 2.67 & 7.72 & 7.53 & 4.76 & 6.72 & 7.70 & 12.39 & 6.17 & 7.15 & 10.92 & 7.39\\
MF-PSN\cite{16_liu2022deep} & \underline{2.07} & 5.83 & 6.88 & 5.00 & \underline{5.90} & 7.46 & 13.38 & 7.20 & 6.81 & 12.20 & 7.27\\
CNN-PS\cite{4_ikehata2018cnn} & 2.12 & \textbf{4.38} & 8.07 & \underline{4.38} & 7.92 & 7.42 & 14.08 & \textbf{5.37} & \underline{6.38} & 12.12 & 7.20\\
PSMF-PSN\cite{38_yang2023accurate} & 2.54 & 5.99 & 7.21 & 5.09 & \textbf{5.52} & 7.75 & \underline{11.40} & 6.91 & \textbf{6.11} & 10.01 & 6.85\\
SR-PSN\cite{27_ju2023estimating} & 2.23 & 5.24 & \underline{6.75} & 4.63 & 6.12 & \underline{7.07} & 12.61 & 5.88 & 6.44 & 10.35 & 6.73\\
NA-PSN\cite{26_ju2022normattention} & 2.93 & \underline{4.65} & 7.12 & 4.65 & 5.99 & 7.49 & 12.28 & 5.96 & 6.42 & 9.93 & 6.72\\
GR-PSN\cite{39_ju2023grpsn} & 2.22 & 5.61 & \textbf{6.73} & \textbf{4.33} & 6.17 & \textbf{6.78} & 12.03 & \underline{5.54} & 6.42 & \underline{9.65} & \underline{6.55}\\
\hline
IGA-PSN(Ours) & 2.20 & 4.83 & 7.08 & 4.63 & 5.92 & 7.25 & \textbf{11.01} & 5.85 & 6.41 & \textbf{9.37} & \textbf{6.46} \\
\hline
\end{tabular}
\label{tab:tab5}
\end{table}
\begin{table}[hbt]
\renewcommand{\arraystretch}{1}
\setlength\tabcolsep{0.3pt}
    \centering
    \caption{benchmark results on our real-world dataset DiLiGenT-$\mathrm{\Pi}$ \cite{25_wang2023diligent}. We calculate the average MAE of the object in each material.}
    \begin{tabular}{|c|cccccc|c|}
    \hline
    material & L2 \cite{1_woodham1980photometric} & PX-Net \cite{28_logothetis2021px} & PS-FCN  \cite{12_chen2018ps} & GPS-Net \cite{50_yao2020gps} & CNN-PS \cite{4_ikehata2018cnn} & NA-PSN \cite{26_ju2022normattention} & Ours \\
    \hline
    Metallic & 7.2 & 9.2 & 5.2 & \underline{5.0} & \textbf{4.9} & \textbf{4.9} & 5.3\\
    Specular & 8.5 & 8.4 & 8.5 & 8.9 & \textbf{7.4} & \underline{7.6} & \textbf{7.4}\\
    Translucent & 17.0 & \textbf{16.0} & 17.3 & 17.6 & \underline{16.8} & 17.0 & 16.9\\
    Rough & 14.6 & 13.7 & 14.5 & 13.8 & \underline{13.5} & \textbf{13.3} & \textbf{13.3}\\
    \hline
    AVG. & 10.5 & 10.8 & 9.8 & 10.1 & \textbf{9.2} & \textbf{9.2} & \underline{9.3} \\
\hline
\end{tabular}
\label{tab:tab6}
\end{table}

\noindent\textbf{Effectiveness of hourglass normal regressor:} Expanding on the advancements in ID(6), we enhance the normal regressor architecture by adding two hourglass-shaped encoder-decoder blocks after the preprocessing module. Results show notable improvements in all metrics for ID(7) compared to ID(6), with a 0.14° reduction in MAE. The addition of stacked hourglass blocks, with three training outputs, significantly enhances the network's learning capabilities.
\subsection{Benchmark comparison}
\textbf{Comparison on DiLiGent Benchmark:} We compare with 9 SOTA methods on public datasets in Table \ref{tab:tab5}. The DiLiGent dataset \cite{11_shi2016benchmark} includes 96 input images for each object, except for the ``Bear" object, where only the latter 76 images were used due to damage in the first 20 images \cite{4_ikehata2018cnn}. Our average MAE is 6.46°, positioning IGA-PSN among top-performing techniques. It notably achieves impressive results on objects with high-frequency regions like ``Buddha",``Harvest" and ``Reading". Further insights are shown in Fig. \ref{fig:img4}, demonstrating our model's precision in reconstructing normals in challenging areas such as pockets in ``Harvest" and clothing in ``Reading" involving intricate shadows, highlights, and reflections. This showcases IGA-PSN's strength to handle complex surface details.

\noindent\textbf{Comparison on DiLiGenT-$\mathrm{\Pi}$ dataset:} 
The reconstruction of shape details is crucial in PS techniques, especially for objects with intricate features like reliefs, badges, and coins in everyday scenarios. To scrutinize our model’s efficacy in reconstructing surfaces with close proximity, we evaluate IGA-PSN's efficacy in reconstructing such surfaces by comparing it with state-of-the-art methods on the DiLiGenT-$\mathrm{\Pi}$ dataset \cite{25_wang2023diligent}. As shown in Table \ref{tab:tab6}, we achieve a MAE of 9.3 on DiLiGenT-$\mathrm{\Pi}$ \cite{25_wang2023diligent}. However, our model's performance on objects with Translucent materials is less pronounced. This is due to limited training data instances for Translucent materials, resulting in challenges in predicting normal vectors accurately for these pixels. Moreover, translucent surfaces pose significant challenges in photometric stereo methods due to the complex dynamics of subsurface scattering, often leading to blurred surface details. Therefore, our model has limitations in reconstructing objects with translucent surfaces.
\section{Conclusion}
In this paper, we propose an image gradient-aided photometric stereo neural network and introduce an attention feature fusion module for adaptive information exchange among features.  We integrate an hourglass regressor with supervision to regularize normal regression, preserving object surface details.  Through ablation studies and comparisons on public datasets, we validate our method's performance, especially in high-frequency regions under spatially varying BRDF surfaces.  The use of gradient guidance is common in image super-resolution tasks, prompting further exploration of its potential in super-resolution PS tasks.

\vspace{1em}\noindent\textbf{Acknowledgement.} This work is being supported by the National Science Fund of China (Grant No. 41927805).
%
%
%

\begin{thebibliography}{8}
\bibitem{1_woodham1980photometric}
R.~J. Woodham, Photometric method for determining surface orientation from multiple images. In: Optical engineering, vol.~19, pp. 139-144 (1980)

\bibitem{2_shi2013bi}
B.~Shi, et~al.: Bi-polynomial modeling of low-frequency reflectances. In: IEEE transactions on pattern analysis and machine intelligence, vol. 36, no. 6, pp. 1078-1091 (2013)

\bibitem{3_santo2017deep}
H.~Santo, et~al.: Deep photometric stereo network. In: Proceedings of the IEEE international conference on computer vision workshops, pp. 501-509 (2017)

\bibitem{4_ikehata2018cnn}
S.~Ikehata, Cnn-ps: Cnn-based photometric stereo for general non-convex surfaces. In: Proceedings of the European conference on computer vision (ECCV) (2018)

\bibitem{6_mukaigawa2007analysis}
Y.~Mukaigawa, et~al.: Analysis of photometric factors based on photometric linearization. In: Journal of The Optical Society of America A-optics Image Science and Vision, vol.~24, no.~10, pp. 3326-3334 (2007)

\bibitem{7_holroyd2008photometric}
M.~Holroyd, et~al.: A photometric approach for estimating normals and tangents. In: ACM Transactions on Graphics (TOG), vol.~27, no.~5, pp. 1-9 (2008)

\bibitem{8_wu2009photometric}
T.-P. Wu, et~al.: Photometric stereo via expectation maximization. In: IEEE transactions on pattern analysis and machine intelligence, vol.32, no.3, pp.546-560 (2009)

\bibitem{11_shi2016benchmark}
B.~Shi, et~al.: A benchmark dataset and evaluation for non-lambertian and uncalibrated photometric stereo. In: Proceedings of the IEEE Conference on Computer Vision and Pattern Recognition, pp. 3707-3716 (2016)

\bibitem{12_chen2018ps}
G.~Chen, et~al.: Ps-fcn: A flexible learning framework for photometric stereo. In: Proceedings of the European conference on computer vision (ECCV), pp. 3-18 (2018)

\bibitem{13_wann2023practical}
H.~Wann~Jensen, et~al.: A practical model for subsurface light transport. In: Seminal Graphics Papers: Pushing the Boundaries, Volume 2, pp. 319-326 (2023)

\bibitem{15_chen2020deep}
G.~Chen, et~al.: Deep photometric stereo for non-lambertian surfaces. In: IEEE Transactions on Pattern Analysis and Machine Intelligence, vol.~44, no.~1, pp. 129-142 (2020)

\bibitem{16_liu2022deep}
Y.~Liu, Y.~Ju, et~al.: A deep-shallow and global--local multi-feature fusion network for photometric stereo. In: Image and Vision Computing, vol. 118, p. 104368, (2022)

\bibitem{17_taniai2018neural}
T.~Taniai, et~al.: Neural inverse rendering for general reflectance photometric stereo. In: International Conference on Machine Learning, pp. 4857-4866 (2018)

\bibitem{18_ikehata2012robust}
S.~Ikehata, et~al.: Robust photometric stereo using sparse regression. In: Proceedings of the IEEE Conference on Computer Vision and Pattern Recognition (2012)

\bibitem{19_ma2020structure}
C.~Ma, et~al.: Structure-preserving super resolution with gradient guidance. In: Proceedings of the IEEE conference on computer vision and pattern recognition (2020)


\bibitem{21_su2021pixel}
Z.~Su, et~al.: Pixel difference networks for efficient edge detection. In: Proceedings of the IEEE/CVF international conference on computer vision, pp. 5117-5127 (2021)

\bibitem{22_tozza2016direct}
S.~Tozza, R.~Mecca, et~al.: Direct differential photometric stereo shape recovery of diffuse and specular surfaces. In: Journal of Mathematical Imaging and Vision, vol.~56, pp. 57-76 (2016)

\bibitem{23_chang2018pyramid}
J. Chang, Y. Chen, et~al.: Pyramid stereo matching network. In:Proceedings of the IEEE conference on computer vision and pattern recognition, pp. 5410-5418 (2018)

\bibitem{24_woo2018cbam}
S.~Woo, J.~Park, et~al.: Cbam: Convolutional block attention module. In: Proceedings of the European conference on computer vision (ECCV), pp. 3-19 (2018)

\bibitem{25_wang2023diligent}
F.~Wang, J.~Ren, et~al.: Diligent-pi: Photometric stereo for planar surfaces with rich details-benchmark dataset and beyond. In: Proceedings of the IEEE/CVF International Conference on Computer Vision, pp. 9477-9487 (2023)

\bibitem{26_ju2022normattention}
Y.~Ju, B.~Shi, M.~Jian, L.~Qi, et~al.: Normattention-psn: A high-frequency region enhanced photometric stereo network with normalized attention. In: International Journal of Computer Vision, vol.130, no.12, pp. 3014-3034 (2022)

\bibitem{27_ju2023estimating}
Y.~Ju, M.~Jian, et~al.: Estimating high-resolution surface normals via low-resolution photometric stereo images. In: IEEE Transactions on Circuits and Systems for Video Technology (2023) 

\bibitem{28_logothetis2021px}
F.~Logothetis, I.~Budvytis, et~al.: Px-net: Simple and efficient pixel-wise training of photometric stereo networks. In Proceedings of the IEEE/CVF International Conference on Computer Vision, pp. 12757-12766, (2021)

\bibitem{29_ju2022deep}
Y.~Ju, et~al.: Deep learning methods for calibrated photometric stereo and beyond. In: IEEE Transactions on Pattern Analysis and Machine Intelligence (2024)

\bibitem{31_ikehata2022ps}
S.~Ikehata, Ps-transformer: Learning sparse photometric stereo network using self-attention mechanism. In: Proceedings of British Machine Vision Conference (2021)

\bibitem{32_zheng2020summary}
Q.~Zheng, et~al.: Summary study of data-driven photometric stereo methods. In: Virtual Reality \& Intelligent Hardware, vol.~2, no.~3, pp. 213-221 (2020)

\bibitem{34_johnson2011shape}
M.~K. Johnson, et~al.: Shape estimation in natural illumination. In: Proceedings of the IEEE Conference on Computer Vision and Pattern Recognition (2011)

\bibitem{35_wiles2017silnet}
A.~Zisserman, et~al.: SilNet: Single- and multi-View reconstruction by learning from silhouettes. In: Proceedings of British Machine Vision Conference (2017)

\bibitem{37_oren1994generalization}
M.~Oren, S.~K. Nayar, et~al.: Generalization of lambert's reflectance model. In: Proceedings of the 21st annual conference on Computer graphics and interactive techniques (1994)

\bibitem{38_yang2023accurate}
Y.~Yang, J.~Liu, Y.~Ni, et~al.:Accurate normal measurement of non-lambertian complex surface based on photometric stereo. In: IEEE Transactions on Instrumentation and Measurement, pp. 1-11 (2023)

\bibitem{39_ju2023grpsn}
Y.~Ju, B.~Shi, Y.~Chen, et~al: Gr-psn: Learning to estimate surface normal and reconstruct photometric stereo images. In: IEEE Transactions on Visualization and Computer Graphics, pp. 1-16 (2023)

\bibitem{41_Isola_2017_CVPR}
P.~Isola, J.-Y. Zhu, et~al.: Image-to-image translation with conditional adversarial networks. In: Proceedings of the IEEE Conference on Computer Vision and Pattern Recognition (2017)

\bibitem{42_ju2023deep}
Y.~Ju, M.~Jian, et~al.: Deep discrete wavelet transform network for photometric stereo. In: International Conference on Digital Signal Processing (2023)

\bibitem{43_book}
Jähne, Bernd. Digital image processing. Springer Science \& Business Media, (2005)

\bibitem{50_yao2020gps}
Z.~Yao, et~al.: Gps-net: Graph-based photometric stereo network. In: Advances in Neural Information Processing Systems, vol.~33, pp. 10306-10316 (2020)




\end{thebibliography}
%

\end{document}